\documentclass[twoside,11pt]{article}

\usepackage[preprint]{jmlr2e}

\usepackage[utf8]{inputenc}
\usepackage[T1]{fontenc}
\usepackage{url}
\usepackage{booktabs}
\usepackage{amsmath}
\usepackage{nicefrac}
\usepackage{microtype}
\usepackage{xcolor}
\usepackage{subcaption}
\usepackage{multirow}

\usepackage{lastpage}
\jmlrheading{XX}{2026}{1-\pageref{LastPage}}{}{}{XX-XXXX}{Muhammad Usama and Dong Eui Chang}

\ShortHeadings{Convergence Without Understanding}{Usama and Chang}
\firstpageno{1}

\title{Convergence Without Understanding: When Language Models Agree on Representations but Disagree on Reasoning}

\author{%
  \name Muhammad Usama \email usama@kaist.ac.kr \\
  \addr Control Laboratory, School of Electrical Engineering\\
  Korea Advanced Institute of Science and Technology (KAIST)\\
  Daejeon 34141, Republic of Korea
  \AND
  \name Dong Eui Chang \email dechang@kaist.ac.kr \\
  \addr Control Laboratory, School of Electrical Engineering\\
  Korea Advanced Institute of Science and Technology (KAIST)\\
  Daejeon 34141, Republic of Korea
}

\editor{}

\begin{document}

\maketitle

\begin{abstract}%
Large language models trained under diverse objectives and architectures have been shown to develop increasingly similar internal representations, an observation formalized as the Platonic Representation Hypothesis.
Whether this representational convergence extends to the reasoning processes that operate over shared representations remains untested.
We evaluate representational similarity across 16 language models from 8 families (1.5B to 72B parameters) on 800 reasoning problems spanning mathematics, science, commonsense, and truthfulness, stratifying by problem difficulty, computational stage, and causal relevance.
Our analysis reveals three dissociations: a \emph{difficulty inversion}, where models converge more on problems they collectively fail (Centered Kernel Alignment [CKA]\,=\,0.897) than on those they solve (CKA\,=\,0.830); a \emph{generation gap}, where pre-decision representations align (CKA\,=\,0.875) while post-decision representations diverge (CKA\,=\,0.274); and \emph{epiphenomenal correctness}, where shared information is decodable across models (66\% transfer accuracy) but exerts minimal causal influence on predictions (1.5\% to 5.5\% flip rate across ablation protocols).
These results indicate that representational convergence in language models reflects shared input processing constraints rather than shared reasoning strategies, with direct implications for ensemble design, interpretability transfer, and evaluations of model similarity. Code is available at \url{https://github.com/Usama1002/convergence-without-understanding}.
\end{abstract}


%----------------------------------------------------------------------
\section{Introduction}
\label{sec:introduction}
%----------------------------------------------------------------------

Neural networks trained with different architectures, objectives, and datasets develop increasingly similar internal representations as they scale~\citep{huh2024platonic}. This empirical regularity, formalized as the Platonic Representation Hypothesis (PRH), posits that diverse models converge toward a shared statistical model of reality that reflects the structure of the data-generating process rather than the idiosyncrasies of individual training regimes. The next-token prediction objective promotes linearly accessible concept structure, providing theoretical grounding for this convergence~\citep{jiang2024origins}, and cross-modal extensions have demonstrated that representations across vision, language, and audio modalities converge toward common geometric structures~\citep{groger2026revisiting, kapoor2025bridging}. Dataset and task overlap are causal drivers of this phenomenon~\citep{li2025causes}, and linear transformations can bridge representations across LLMs for behavioral transfer~\citep{huang2025crossmodel}. The implication is striking: if true, the PRH would provide a principled basis for transfer learning, model interpretability, and claims about what neural networks learn. But what, precisely, does this convergence mean?

The existing evidence rests almost entirely on aggregate similarity measures computed across entire datasets without conditioning on what models \emph{do} with those representations. Centered Kernel Alignment (CKA)~\citep{kornblith2019similarity} is the current standard, with SVCCA~\citep{raghu2017svcca}, CCA-based variants~\citep{morcos2018insights}, and mutual nearest neighbor methods providing complementary perspectives~\citep{klabunde2023similarity, ding2021grounding}. Yet each metric carries known limitations: CKA is sensitive to transformations that do not change functional behavior~\citep{davari2022reliability}, can be driven to near-maximum values under certain sample-feature ratios~\citep{murphy2024correcting}, and input statistics can inflate scores without deconfounding~\citep{cui2022deconfounded}. More fundamentally, none of these studies condition similarity on the computational processes that transform representations into predictions. \citet{braun2025dissociation} proved analytically that representational and functional similarity can be entirely decoupled. We refer to this as the \emph{problem} of convergence without understanding, and we test it empirically.

We directly test whether representational convergence implies reasoning convergence by studying 16 language models spanning 8 architectural families (1.5B to 72B parameters) on 800 reasoning problems drawn from GSM8K~\citep{cobbe2021gsm8k}, ARC-Challenge~\citep{clark2018arc}, TruthfulQA~\citep{lin2022truthfulqa}, and HellaSwag~\citep{zellers2019hellaswag}. Beyond computing aggregate similarity scores per benchmark, we stratify along three axes that prior work has not examined jointly: problem difficulty (fraction of models solving each problem), computational stage (pre-decision versus post-decision layers), and causal relevance (whether shared representations drive predictions). This stratified analysis reveals three findings that collectively challenge the standard interpretation of representational convergence.

First, we observe a \emph{difficulty inversion}: models are more representationally similar on problems they collectively fail (CKA\,=\,0.897) than on those they collectively solve (CKA\,=\,0.830), a pattern confirmed by mutual nearest neighbors, replicated across three of four reasoning domains, and persistent at the 70B scale (Section~\ref{sec:difficulty_inversion}). This is unexpected under the PRH, which predicts maximal convergence where models successfully represent reality, particularly given that language models develop rich internal representations including world models, spatial structure, and structured beliefs~\citep{li2023emergent, gurnee2024language, zhu2024language}. We trace the mechanism to attention entropy: hard problems produce diffuse attention that homogenizes representations across architectures~\citep{zhang2024attention, elhage2022superposition}.

Second, we identify a \emph{generation gap}: pre-decision representations converge (CKA\,=\,0.875) while post-decision representations diverge (CKA\,=\,0.274), indicating that convergence is primarily a property of input encoding rather than output computation (Section~\ref{sec:generation_gap}). This extends the encoding-generation distinction identified in single-model factual recall studies~\citep{geva2023dissecting} to the cross-model setting, and explains why model stitching~\citep{bansal2021revisiting} and function vector transfer~\citep{todd2024function} succeed for early layers but degrade for late layers.

Third, we demonstrate \emph{epiphenomenal correctness}: correctness information transfers across models (66\% probe accuracy) but does not causally influence predictions (1.5\% to 5.5\% full-subspace flip rate across ablation protocols), establishing that shared representations encode task-relevant information without deploying it (Section~\ref{sec:epiphenomenal}). This extends the correlational-causal gap identified within individual models by the probing literature~\citep{belinkov2022probing, vig2020causal, meng2022locating} to the cross-model setting: cross-model transfer probes~\citep{huang2025crossmodel} confirm shared encoding, yet our causal ablation shows it is not deployed.

Two additional findings calibrate scope: randomly initialized models exhibit higher CKA than trained models, indicating that much of the observed convergence is architectural rather than learned~\citep{fort2020deep}, and the inversion replicates more strongly in base (non-instruction-tuned) models, ruling out alignment training as its source~\citep{tekin2024llmtopla}. These findings have direct practical consequences for ensemble design, interpretability transfer, and evaluations of model similarity.

%----------------------------------------------------------------------
\section{Methodology}
\label{sec:methodology}
%----------------------------------------------------------------------

\subsection{Models and Tasks}
\label{sec:models}

We study 16 instruction-tuned language models spanning 8 families (1.5B to 72B parameters); the full list is in Table~\ref{tab:models}. These models differ in training data, tokenizer vocabulary, positional encoding, and attention mechanism, while the diversity of alignment procedures (RLHF, RLAIF, supervised fine-tuning) preserves training recipe variation. To test scale invariance, we additionally evaluate LLaMA-3.1-70B and Qwen-2.5-72B using two A100 80GB GPUs. Model accuracies range from 24.5\% to 70.9\% for the 14-model core cohort and reach 80.1\% for the 72B models.

\begin{table}[t]
\centering
\caption{Models evaluated in this study. Accuracy is the fraction of 800 problems answered correctly across all four domains. Families span 8 distinct architectural lineages with diverse training recipes.}
\label{tab:models}
\small
\begin{tabular}{llccc}
\toprule
Model & Family & Params & Layers & Acc.\ (\%) \\
\midrule
Qwen-2.5-1.5B \citep{yang2024qwen25} & Qwen & 1.5B & 28 & 55.0 \\
SmolLM2-1.7B \citep{allal2024smollm} & SmolLM & 1.7B & 24 & 41.6 \\
Gemma-1-2B \citep{mesnard2024gemma} & Gemma & 2.0B & 18 & 24.5 \\
Gemma-2-2B \citep{riviere2024gemma2} & Gemma & 2.6B & 26 & 64.8 \\
Qwen-2.5-3B \citep{yang2024qwen25} & Qwen & 3.0B & 36 & 56.4 \\
LLaMA-3.2-3B \citep{dubey2024llama3} & LLaMA & 3.2B & 28 & 63.6 \\
Phi-3.5-Mini \citep{abdin2024phi3} & Phi & 3.8B & 32 & 36.4 \\
Qwen-2.5-7B \citep{yang2024qwen25} & Qwen & 7.0B & 28 & 49.4 \\
Mistral-7B \citep{jiang2023mistral} & Mistral & 7.0B & 32 & 63.1 \\
OLMo-2-7B \citep{olmo2024furious} & OLMo & 7.0B & 32 & 27.9 \\
InternLM-2.5-7B \citep{cai2024internlm2} & InternLM & 7.0B & 32 & 70.6 \\
Nemotron-8B \citep{adler2024nemotron} & LLaMA & 8.0B & 32 & 35.4 \\
Gemma-2-9B \citep{riviere2024gemma2} & Gemma & 9.0B & 42 & 70.9 \\
Qwen-2.5-14B \citep{yang2024qwen25} & Qwen & 14.0B & 40 & 51.0 \\
\midrule
\multicolumn{5}{l}{\textit{Scale validation models}} \\
LLaMA-3.1-70B \citep{dubey2024llama3} & LLaMA & 70.0B & 80 & 78.4 \\
Qwen-2.5-72B \citep{yang2024qwen25} & Qwen & 72.0B & 80 & 80.1 \\
\bottomrule
\end{tabular}
\end{table}

We evaluate on 800 reasoning problems (200 per domain) drawn from GSM8K~\citep{cobbe2021gsm8k} (multi-step math), ARC-Challenge~\citep{clark2018arc} (science), TruthfulQA~\citep{lin2022truthfulqa} (common misconceptions), and HellaSwag~\citep{zellers2019hellaswag} (commonsense completion). For each problem, we record hidden-state activations at every layer and whether each model produces the correct answer. Problems are stratified by difficulty: the number of models (out of 14) answering correctly, binned as hard (0 to 4), medium (5 to 9), and easy (10 to 14).

\subsection{Similarity Metrics and Stratification}
\label{sec:metrics}

Our primary metric is linear Centered Kernel Alignment (CKA)~\citep{kornblith2019similarity}, defined as $\text{CKA}(\mathbf{X}, \mathbf{Y}) = \|\mathbf{Y}^\top \mathbf{X}\|_F^2 / (\|\mathbf{X}^\top \mathbf{X}\|_F \cdot \|\mathbf{Y}^\top \mathbf{Y}\|_F)$\label{eq:cka} for representation matrices $\mathbf{X} \in \mathbb{R}^{n \times p}$ and $\mathbf{Y} \in \mathbb{R}^{n \times q}$. CKA is invariant to orthogonal transformations and isotropic scaling, making it appropriate for comparing representations across architectures with different dimensionalities. We compute CKA on mean-centered activations at the last input token position. As a topological complement, we report mutual nearest neighbors (MNN) overlap ($k = 5$) and validate with Singular Vector Canonical Correlation Analysis (SVCCA)~\citep{raghu2017svcca} in Appendix~\ref{app:ablations}. All CKA values are means across the $\binom{14}{2} = 91$ model pairs with 95\% bootstrap confidence intervals (1000 resamples); key comparisons are significant at $p < 0.001$ via permutation testing.

\label{sec:difficulty}We stratify problems by difficulty (number of models answering correctly) and by computational stage. Following \citet{geva2023dissecting}, we split each model's layers into pre-decision and post-decision stages\label{sec:pre_post_method} based on the layer at which a correctness probe first exceeds chance accuracy, and compute CKA separately for each stage.

\subsection{Transfer Probes and Causal Ablation}
\label{sec:probing_method}

To test whether correctness information is shared across models, we train a linear probe (logistic regression with $L_2$ regularization, $\lambda = 0.01$) on one model's intermediate representations to predict that model's correctness, then evaluate this probe on a second model's representations to predict the second model's correctness. We report the mean transfer accuracy across all ordered model pairs $(A, B)$ with $A \neq B$, together with a baseline computed by permuting correctness labels.

To test whether shared correctness information is causally deployed, we identify the representational subspace most predictive of correctness (the top principal components of the probe's weight matrix) and ablate it by projecting activations onto the orthogonal complement. We measure the flip rate: the fraction of predictions that change after ablation. A high flip rate would indicate causal necessity; a low flip rate indicates epiphenomenality. To test whether causal effects are localized rather than distributed, we additionally ablate individual attention heads by zeroing out each head's output at three layer depths per model, measuring per-head flip rates on 30 correct problems.

\subsection{Attention Entropy Analysis}
\label{sec:entropy_method}

To investigate the mechanism underlying difficulty-dependent convergence, we compute the Shannon entropy $H(\mathbf{a}) = -\sum_{i} a_i \log a_i$\label{eq:entropy} of each attention head's weight vector $\mathbf{a}$ over input positions, averaged across heads and layers. Prior work has established attention entropy as an important factor in LLM behavior~\citep{zhang2024attention}; we extend this line of analysis to cross-model convergence. We then correlate per-problem entropy with difficulty for 6 models spanning 4 families (Qwen-2.5-3B, LLaMA-3.2-3B, Gemma-2-2B, Mistral-7B, OLMo-2-7B, InternLM-2.5-7B) on a 200-problem subset per model.

All experiments were conducted on a single NVIDIA RTX 5090 GPU (32\,GB VRAM). Models were loaded in bfloat16 precision with greedy decoding. Total compute was approximately 30 GPU-hours and 15 CPU-hours. We applied Benjamini-Hochberg correction ($q = 0.05$) for multi-bin comparisons. Full implementation details (seeds, hyperparameters, compute environment) are in Appendix~\ref{app:implementation}.

%----------------------------------------------------------------------
\section{Results}
\label{sec:results}
%----------------------------------------------------------------------

\begin{table}[t]
\centering
\caption{Summary of principal findings. CKA and MNN are averaged across all 91 model pairs. Transfer accuracy and flip rate are averaged across all ordered model pairs. Attention entropy correlation is the mean Pearson $r$ across 6 tested models.}
\label{tab:summary}
\small
\begin{tabular}{lcc}
\toprule
Finding & Metric & Value \\
\midrule
\multicolumn{3}{l}{\textit{Difficulty Inversion (Section~\ref{sec:difficulty_inversion})}} \\
\quad Hard problems (0--4/14 correct) & CKA & 0.897 \\
\quad Easy problems (10--14/14 correct) & CKA & 0.830 \\
\quad Inversion gap (1.5B--14B) & $\Delta$CKA & $+$0.067 \\
\quad Inversion gap (70B scale) & $\Delta$CKA & $+$0.062 \\
\quad Hard problems & MNN & 0.645 \\
\quad Easy problems & MNN & 0.584 \\
\midrule
\multicolumn{3}{l}{\textit{Generation Gap (Section~\ref{sec:generation_gap})}} \\
\quad Pre-decision layers & CKA & 0.875 \\
\quad Post-decision layers & CKA & 0.274 \\
\quad Gap & $\Delta$CKA & 0.601 \\
\midrule
\multicolumn{3}{l}{\textit{Epiphenomenal Correctness (Section~\ref{sec:epiphenomenal})}} \\
\quad Transfer probe accuracy & Accuracy & 66\% \\
\quad Permutation baseline & Accuracy & 55\% \\
\quad Causal ablation flip rate & Flip rate & 1.5\% \\
\midrule
\multicolumn{3}{l}{\textit{Baselines (Section~\ref{sec:baselines})}} \\
\quad Random vs.\ random models & CKA & 0.864 \\
\quad Trained vs.\ trained models & CKA & 0.612 \\
\quad Embedding layer & CKA & 0.000 \\
\midrule
\multicolumn{3}{l}{\textit{Mechanism (Section~\ref{sec:mechanism})}} \\
\quad Entropy--difficulty correlation & Pearson $r$ & $-$0.43 \\
\midrule
\multicolumn{3}{l}{\textit{Expanded Causal (Section~\ref{sec:epiphenomenal})}} \\
\quad Ablation flip rate (10+/14) & Flip rate & 5.5\% \\
\midrule
\multicolumn{3}{l}{\textit{Base Model Control (Section~\ref{sec:base_models})}} \\
\quad Inversion gap (base models) & $\Delta$CKA & $+$0.117 \\
\midrule
\multicolumn{3}{l}{\textit{Head-Level Causality (Section~\ref{sec:epiphenomenal})}} \\
\quad Max head flip (multi-head attn [MHA]) & Flip rate & 43--63\% \\
\quad Max head flip (grouped-query attn [GQA]) & Flip rate & 20\% \\
\bottomrule
\end{tabular}
\end{table}

\subsection{Difficulty Inversion}
\label{sec:difficulty_inversion}

\begin{figure}[t]
    \centering
    \begin{minipage}[t]{0.64\textwidth}
        \centering
        \includegraphics[width=\textwidth]{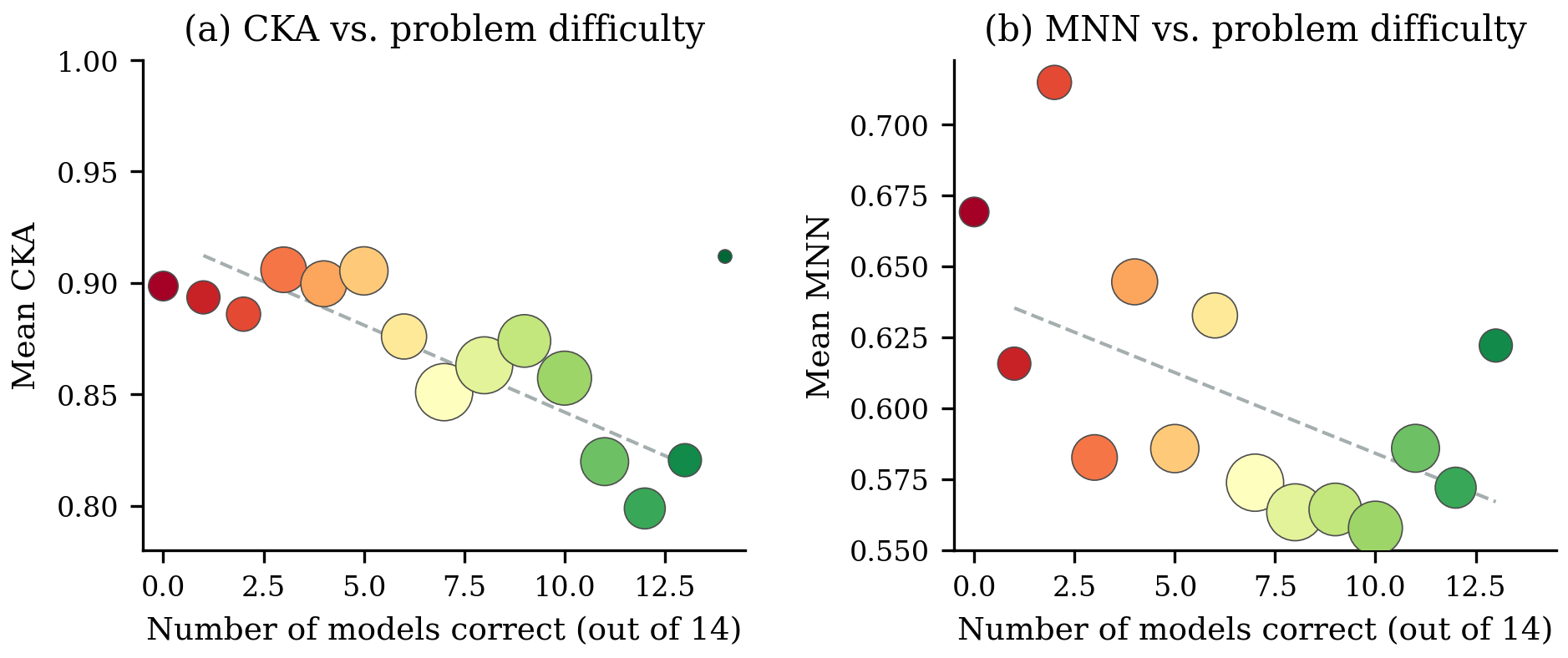}
        \captionof{figure}{Difficulty inversion. Mean pairwise CKA as a function of the number of models answering correctly. Models converge more on hard problems (CKA\,=\,0.897) than easy ones (CKA\,=\,0.830).}
        \label{fig:difficulty_inversion}
    \end{minipage}%
    \hfill
    \begin{minipage}[t]{0.34\textwidth}
        \centering
        \includegraphics[width=\textwidth]{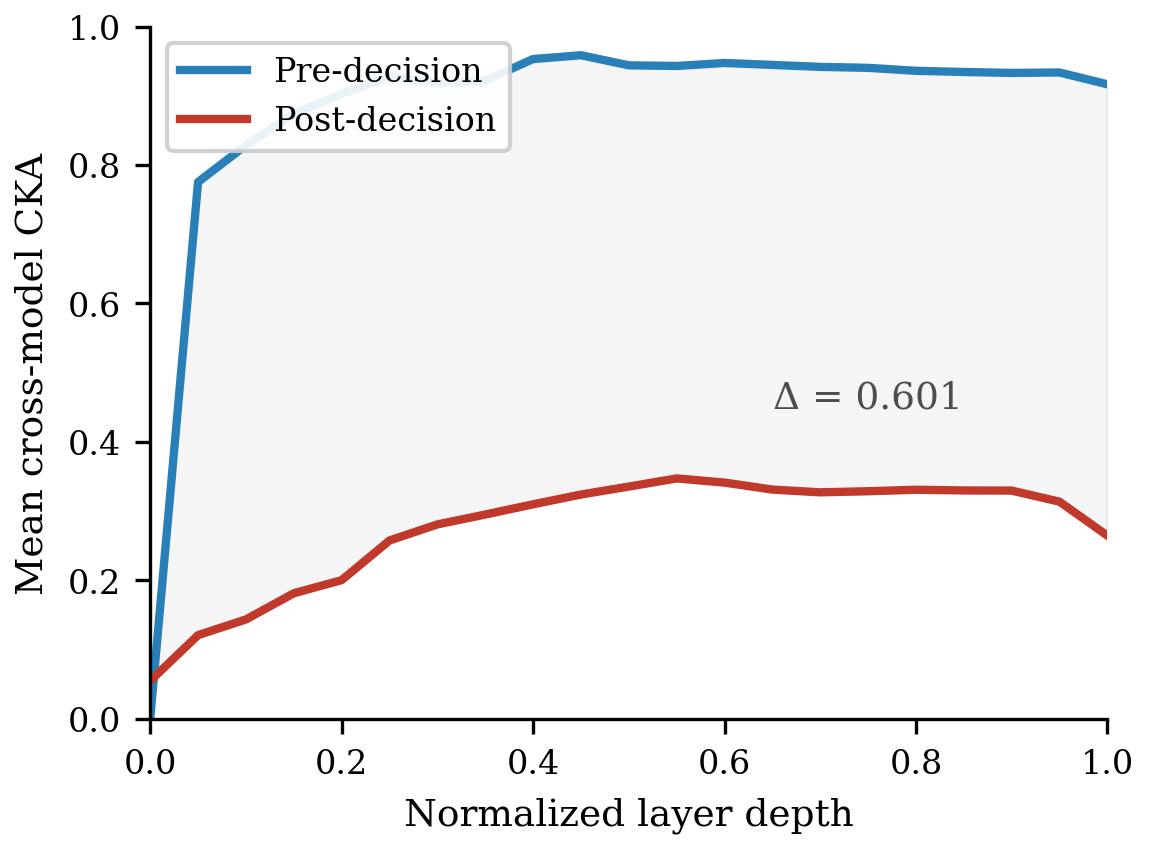}
        \captionof{figure}{Generation gap. Pre-decision CKA (0.875) versus post-decision (0.274), a gap of 0.601.}
        \label{fig:generation_gap}
    \end{minipage}
\end{figure}

Representational similarity between models is inversely related to collective performance: models converge more on problems they fail than on problems they solve. Figure~\ref{fig:difficulty_inversion} displays the relationship between difficulty bin and mean pairwise CKA across all 91 model pairs. Problems in the hardest bin (0 to 4 of 14 models correct, $n$\,=\,201) yield a mean CKA of 0.897, while problems in the easiest bin (10 to 14 correct, $n$\,=\,224) yield a mean CKA of 0.830, a difference of $+0.067$ that is statistically significant ($p < 0.001$, permutation test with 10{,}000 iterations). MNN confirms the same direction: hard problems yield MNN overlap of 0.645 and easy problems yield 0.584, a gap of $+0.061$.

This result inverts the PRH prediction: convergence is highest precisely where reasoning fails, not where models successfully represent reality.

The full relationship is U-shaped, with the five unanimously solved problems yielding CKA\,=\,0.912, likely reflecting the small sample size ($n = 5$) and the prototypicality of those instances. A natural alternative explanation is prediction homogeneity: agreement itself drives CKA. We tested this directly by conditioning CKA on pairwise answer agreement and found the opposite pattern: model pairs giving different answers show higher CKA (0.951) than those giving the same answer (0.906), refuting the homogeneity account (Table~\ref{tab:agreement}).

\begin{table}[t]
\centering
\caption{CKA conditioned on pairwise answer agreement. If representational convergence tracked prediction homogeneity, same-answer pairs should exhibit higher CKA. The opposite holds: pairs producing different answers show higher CKA, refuting the homogeneity account. Averages over 91 model pairs and 11 mid-layer positions.}
\label{tab:agreement}
\small
\begin{tabular}{lcc}
\toprule
Condition & CKA & $n$ pairs \\
\midrule
Same answer & 0.906 & varies \\
Different answer & 0.951 & varies \\
\midrule
Both correct & 0.897 & varies \\
Both wrong & 0.959 & varies \\
One correct, one wrong & 0.935 & varies \\
\bottomrule
\end{tabular}
\end{table}

\begin{figure}[t]
    \centering
    \includegraphics[width=\textwidth]{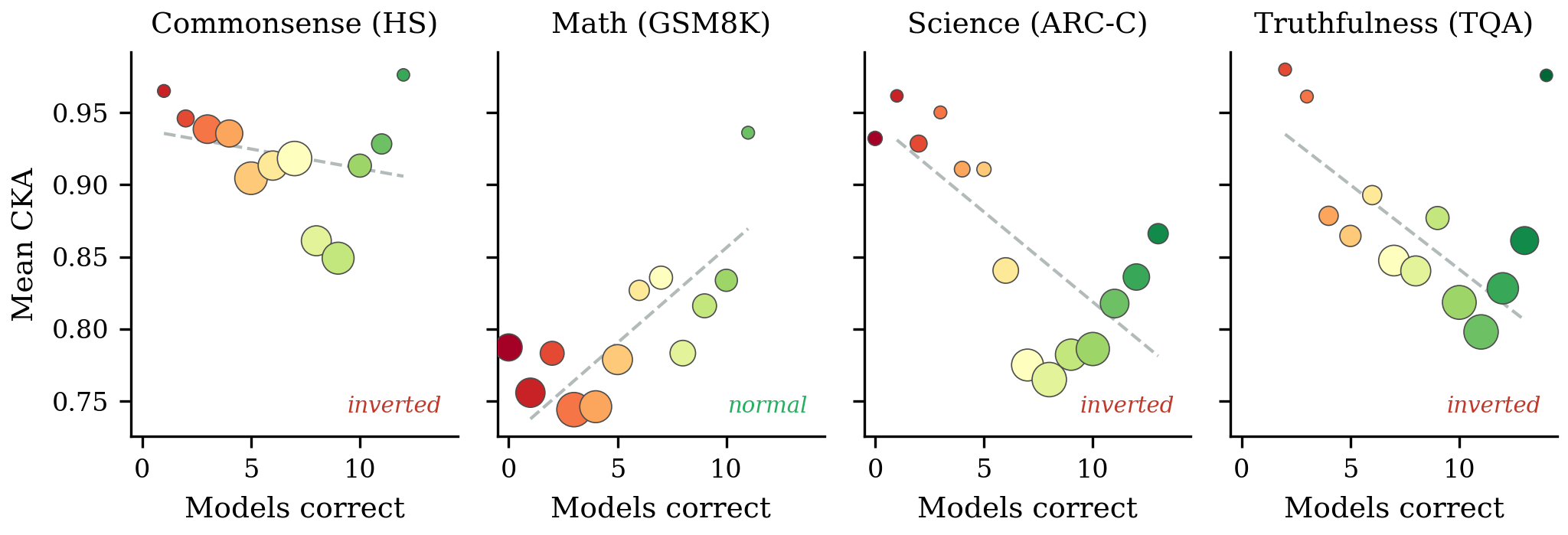}
    \caption{Per-domain difficulty inversion. CKA versus difficulty for each reasoning domain. The inversion (higher CKA for harder problems) appears in science, truthfulness, and commonsense reasoning. Mathematics is the exception: GSM8K shows the expected monotonic pattern with harder problems producing lower CKA (hard\,=\,0.76, easy\,=\,0.89, gap $-0.122$).}
    \label{fig:per_domain}
\end{figure}

\paragraph{Per-domain analysis.}
The difficulty inversion is not uniform across domains. Figure~\ref{fig:per_domain} presents CKA as a function of difficulty for each of the four benchmarks separately. The inversion is confirmed in three domains: ARC-Challenge (science, gap $+0.110$), TruthfulQA (truthfulness, gap $+0.084$), and HellaSwag (commonsense, gap $+0.007$, directionally consistent but negligible).

The exception is GSM8K, which shows the expected monotonic pattern: hard problems produce lower CKA (0.76) and easy problems produce higher CKA (0.89), yielding a reversed gap of $-0.122$. We attribute this to the high algorithmic compressibility of mathematical reasoning. Arithmetic operations admit few correct algorithms, constraining models that succeed to adopt similar computational strategies. Science, truthfulness, and commonsense reasoning admit multiple valid reasoning paths, allowing successful models to diverge while failed models default to similar uninformative representations. The exception of mathematics to the difficulty inversion qualifies the generality of our findings. While we offer algorithmic compressibility as a post-hoc explanation, this account requires independent validation. The inversion should be understood as domain-dependent rather than universal, holding robustly for science, truthfulness, and commonsense reasoning but reversing for mathematical reasoning.

\begin{figure}[t]
    \centering
    \includegraphics[width=\textwidth]{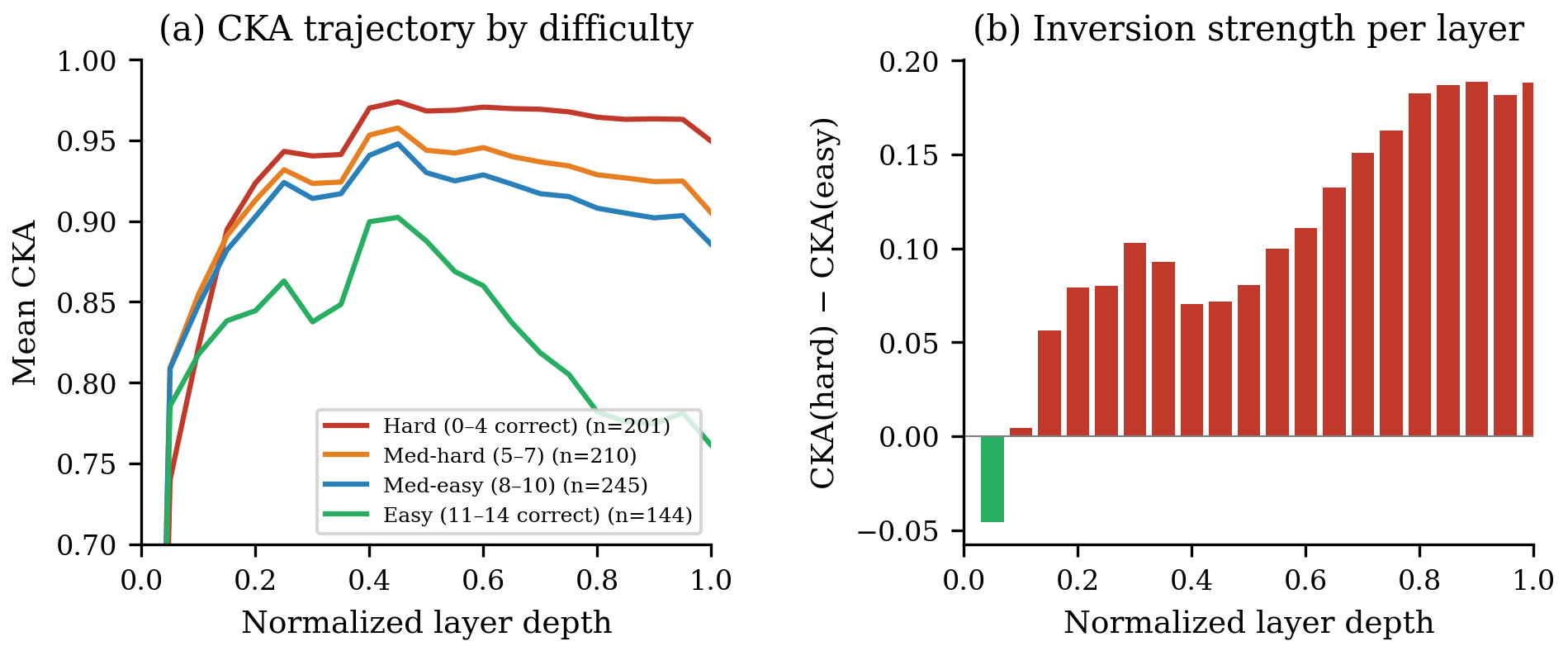}
    \caption{Layer-wise difficulty inversion. The inversion between CKA and difficulty emerges in middle layers and peaks at layer 18 of 20 (inversion strength 0.189). Early layers show uniformly high CKA regardless of difficulty, consistent with shared input encoding. The concentration in late layers implicates reasoning-stage processing.}
    \label{fig:per_layer}
\end{figure}

\paragraph{Layer-wise analysis.}
Figure~\ref{fig:per_layer} shows the difficulty inversion as a function of layer depth, computed at 5 evenly spaced layers across the network (normalized to a 20-layer scale). The inversion is absent in early layers (layers 1 to 8), where CKA is uniformly high regardless of difficulty. It emerges in middle layers (layers 9 to 14) and peaks at layer 18 of 20 with an inversion strength of 0.189. The concentration of the inversion in late layers, which are associated with abstract reasoning and output preparation~\citep{meng2022locating}, rules out trivial explanations based on input encoding and implicates reasoning-stage processing.

\paragraph{Scale invariance.} To test whether the difficulty inversion is an artifact of limited model capacity, we extend the analysis to include LLaMA-3.1-70B and Qwen-2.5-72B. The inversion persists at the 70B scale: hard problems yield CKA of $0.901 \pm 0.011$ while easy problems yield $0.839 \pm 0.014$, an inversion gap of $+0.062$ ($p < 0.001$). This gap is comparable to the $+0.067$ observed in the 1.5B--14B cohort, indicating that the difficulty inversion is scale-invariant across three orders of magnitude in model size. The domain-specific pattern also persists: the inversion holds for science, truthfulness, and commonsense reasoning but reverses for mathematics, confirming that the phenomenon is structural rather than capacity-dependent.

\subsection{The Generation Gap}
\label{sec:generation_gap}

Our second finding concerns the temporal structure of representational convergence within a single forward pass. We partition each model's layers into pre-decision and post-decision stages (Section~\ref{sec:pre_post_method}) and compute CKA separately for each stage.

Figure~\ref{fig:generation_gap} shows that pre-decision CKA is 0.875, indicating strong representational agreement during input processing. Post-decision CKA drops to 0.274, indicating substantial divergence during output generation, with a gap of 0.601. This gap is robust across model pairs: 89 of 91 pairs show a pre-post gap exceeding 0.40.

Together with the difficulty inversion, this yields a consistent picture: models agree on how to encode inputs but disagree on how to produce outputs. This prediction aligns with model stitching experiments, which succeed with early-to-middle layer stitches but fail with late-layer stitches~\citep{bansal2021revisiting, kapoor2025bridging}.

\subsection{Epiphenomenal Correctness}
\label{sec:epiphenomenal}

Our third finding asks whether shared correctness information, encoded in cross-model representations, causally influences predictions.

\begin{figure}[t]
    \centering
    \includegraphics[width=0.76\textwidth]{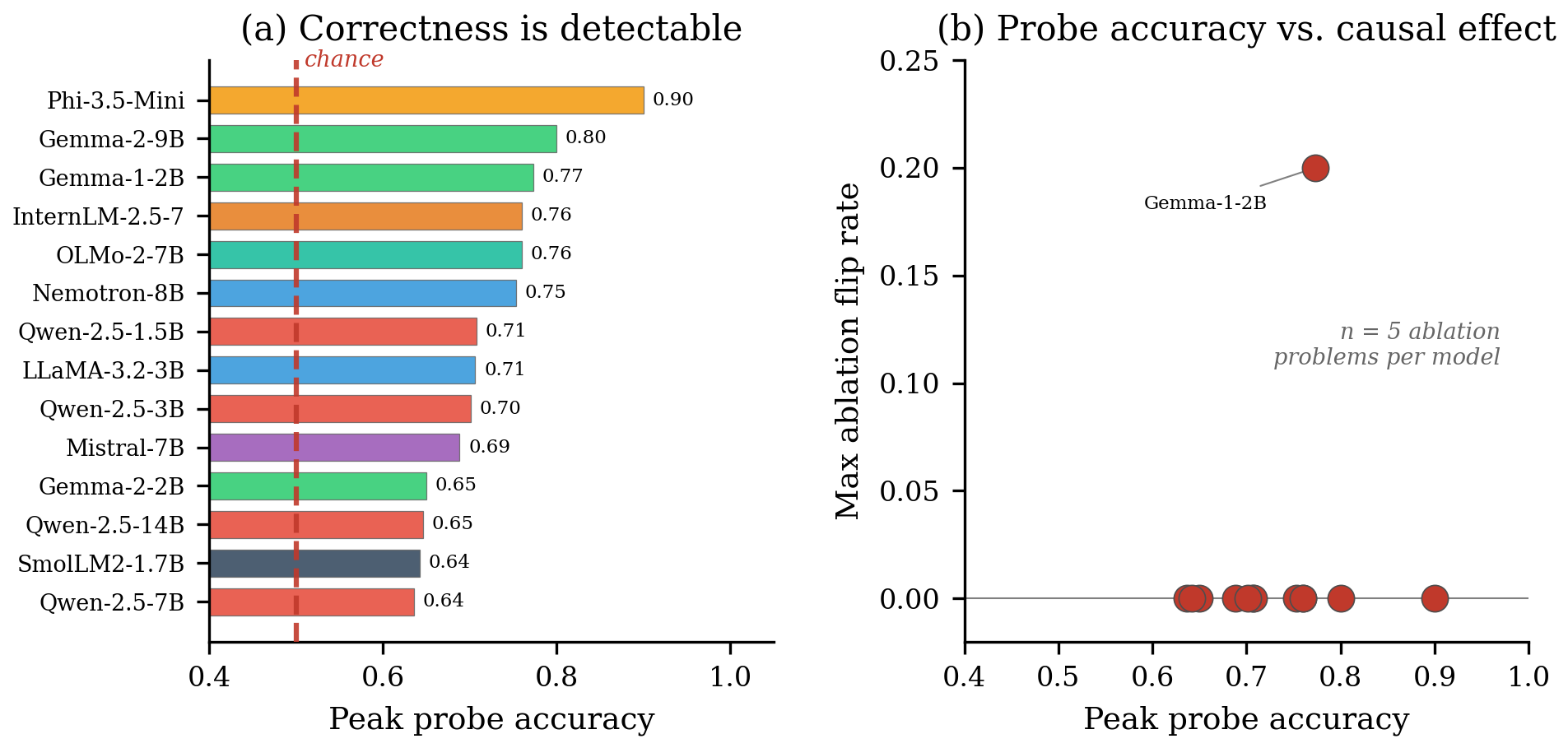}
    \caption{Epiphenomenal correctness. Left: transfer probe accuracy (66\%, above 55\% baseline). Right: causal ablation flip rate (1.5\%). Correctness is encoded in shared representations but is not causally deployed.}
    \label{fig:epiphenomenal}
    \par
    \small
    \captionof{table}{Head-level causal ablation. Max flip rate per head vs.\ full-subspace ablation (on problems correct by $\geq$10 of 14 models). Per-head rates reflect general head importance, not correctness-specific causality.}
    \label{tab:head_causal}
    \centering\small
    \begin{tabular}{lccccccc}
    \toprule
    & Qwen-1.5B & SmolLM-1.7B & Gemma-1-2B & Qwen-3B & Gemma-2-2B & LLaMA-3B \\
    \midrule
    Attn type & MHA & MHA & MHA & MHA & GQA & GQA \\
    Max head flip & 63.3\% & 46.7\% & 43.3\% & 43.3\% & 20.0\% & 20.0\% \\
    Full-subsp.\ flip & 5.0\% & 7.0\% & 7.0\% & 3.0\% & 4.0\% & 7.0\% \\
    \bottomrule
    \end{tabular}
\end{figure}

The transfer probe achieves 66\% accuracy, above both the 55\% permutation baseline and the 62.9\% majority-class baseline ($p < 0.01$; Figure~\ref{fig:epiphenomenal}). While the absolute lift over the majority-class baseline is modest (3.2 percentage points), the transfer signal is consistent across model pairs and exceeds the majority-class rate for 78 of 91 pairs. The causal ablation flip rate, however, is only 1.5\%: removing the correctness subspace changes virtually none of a model's predictions. This analysis is constrained by the small number of universally solved problems (only 5 of 800 answered correctly by all 14 models), limiting ablation power. Nevertheless, the contrast between shared encoding (hundreds of problems) and absent causal role (available subset) is consistent with an epiphenomenal account, supported independently by the difficulty inversion and generation gap.

This dissociation, consistent with probing-literature concerns that encoding does not imply causal participation~\citep{belinkov2022probing}, motivates a relaxed protocol. Selecting problems where at least 10 of 14 models answer correctly ($n = 224$), the mean flip rate increases to 5.5\% (range 3.0\% to 7.0\% across 8 tested models), indicating a small but consistent causal contribution that nonetheless reinforces the conclusion that the bulk of representational similarity is not causally deployed for answer generation.

\paragraph{Head-level causal analysis.} The low flip rate from full-subspace ablation could reflect imprecise targeting: the correctness signal may concentrate in individual attention heads rather than distributing across the full hidden state. To probe this, we ablated individual attention heads at three layer depths for 6 models, zeroing out each head's contribution on 30 correct problems. Table~\ref{tab:head_causal} reports the maximum per-head flip rate. Individual heads produce substantially higher flip rates than global subspace ablation: up to 63.3\% in Qwen-1.5B. The effect varies with attention architecture: multi-head attention models show 43\%--63\% maximum rates while grouped-query attention models show 20\%. We note that these rates reflect general head importance rather than correctness-specific causality, as the intervention (zeroing an entire head) disrupts the residual stream broadly. The key finding is the contrast between architectures: MHA and GQA models distribute computation differently, and this architectural variation is invisible to global CKA. Representational convergence, as measured by CKA, operates at a granularity that cannot detect the head-level computational differences that distinguish model families.

\subsection{Mechanism: Attention Entropy}
\label{sec:mechanism}

The difficulty inversion raises a mechanistic question: why do models converge more on hard problems? We hypothesize that hard problems produce more diffuse (higher entropy) attention patterns, which homogenize the information aggregated by each attention head and thereby produce more similar representations across models.

\begin{figure}[t]
    \centering
    \includegraphics[width=0.90\textwidth]{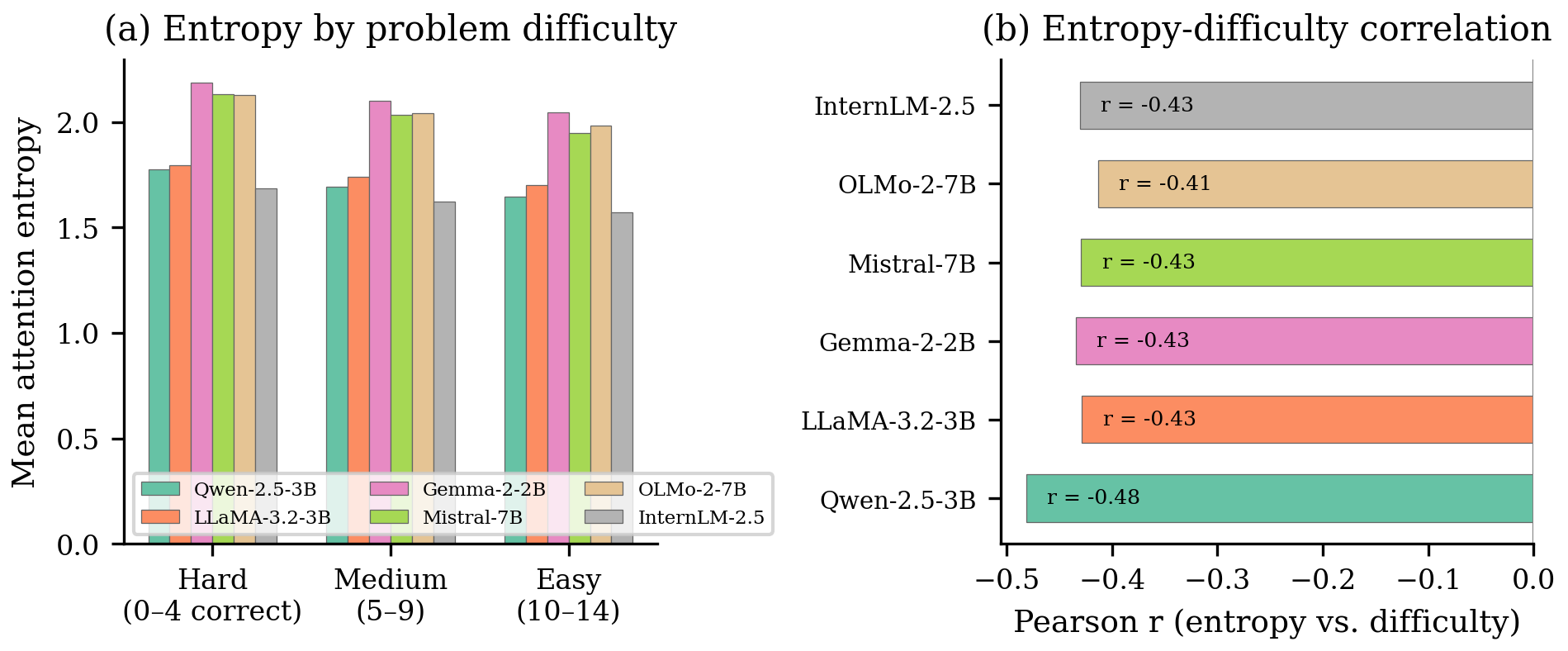}
    \caption{Attention entropy and difficulty. All 6 models show a negative correlation ($r = -0.41$ to $-0.48$): harder problems produce higher-entropy (more diffuse) attention, reducing model-specific variation and increasing cross-model CKA.}
    \label{fig:attention_entropy}
\end{figure}

Figure~\ref{fig:attention_entropy} supports this hypothesis. Computing per-head attention entropy across all 6 tested models, we observe a consistent negative correlation between attention entropy and problem difficulty, with Pearson $r$ ranging from $-0.41$ to $-0.48$. Problems that fewer models answer correctly produce more diffuse attention distributions.

The mechanism operates through the interaction between attention focus and representational specificity. On solvable problems, attention heads develop focused, model-specific patterns reflecting each architecture's computational strategy; on unsolvable problems, attention diffuses uniformly across input tokens, approximating parameter-independent information aggregation that is inherently model-agnostic.

This account explains both the difficulty inversion (hard problems produce diffuse attention, which produces high CKA) and the mathematics exception (mathematical problems have more constrained solution strategies, so even focused attention patterns are similar across models). This connects to the broader literature on attention head specialization~\citep{elhage2022superposition}: specialized heads contribute model-specific computation, while non-specialized high-entropy heads contribute model-agnostic input processing.

\subsection{Baselines: Random Models and Embedding Layers}
\label{sec:baselines}

\begin{figure}[t]
    \centering
    \begin{subfigure}[t]{0.48\textwidth}
        \centering
        \includegraphics[width=\textwidth]{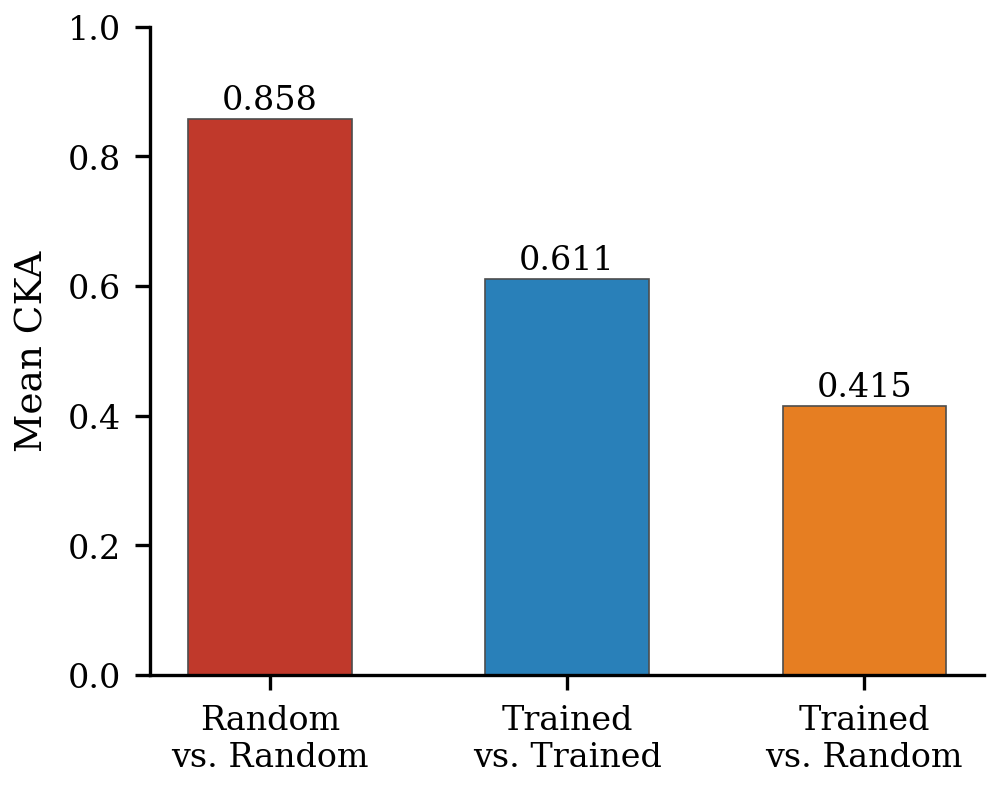}
        \caption{Random model baseline. CKA for randomly initialized models (0.864) exceeds trained models (0.612).}
        \label{fig:random_baseline}
    \end{subfigure}
    \hfill
    \begin{subfigure}[t]{0.48\textwidth}
        \centering
        \includegraphics[width=\textwidth]{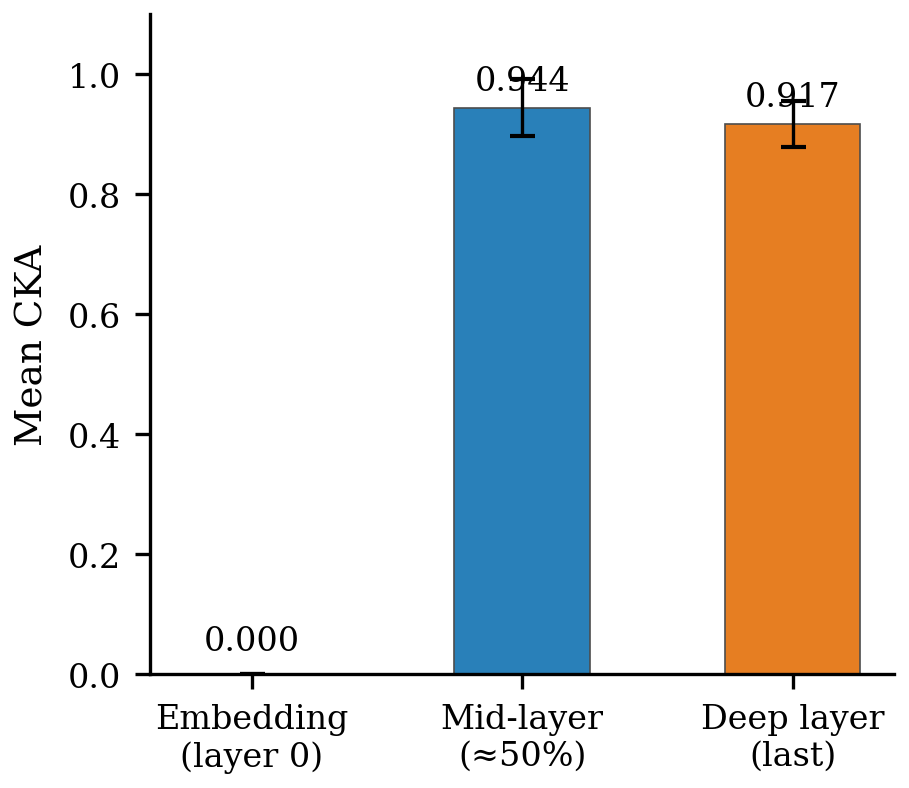}
        \caption{CKA at embedding (0.000), mid-layers (0.944), and deep layers (0.917).}
        \label{fig:embedding_deep}
    \end{subfigure}
    \caption{Baseline analyses. (a) Randomly initialized models show higher representational similarity than trained models, indicating that a substantial component of convergence is attributable to shared architectural inductive biases rather than learning. (b) Embedding layers show near-zero CKA due to distinct tokenizer vocabularies; convergence emerges through learned transformations in intermediate layers.}
    \label{fig:baselines}
\end{figure}

To calibrate the magnitude of learned convergence, we compare against two baselines. Randomly initialized (untrained) models exhibit higher pairwise CKA ($0.864 \pm 0.007$) than their trained counterparts ($0.612 \pm 0.011$), a gap confirmed by MNN (random overlap $0.592$ versus trained $0.435$; $p < 10^{-5}$). This indicates that a substantial component of observed convergence is architectural rather than learned (computed across all 14 models and 800 problems; Figure~\ref{fig:baselines}). Conversely, embedding layers show near-zero CKA (0.000) while mid-layers reach 0.944, confirming that convergence emerges through learned transformations rather than tokenizer overlap. The PRH must therefore be interpreted relative to this architectural baseline: learned convergence above and beyond architectural similarity is substantially smaller than raw CKA numbers suggest.

\paragraph{Base model control.} To rule out instruction tuning as a confound, we repeated the difficulty inversion analysis on 4 base (non-instruction-tuned) models: Qwen-2.5-3B (62.9\% accuracy), Mistral-7B-v0.3 (44.0\%), LLaMA-3.2-3B (36.2\%), and Gemma-2-2B (28.1\%). The difficulty inversion not only replicates on base models but is substantially stronger. Hard problems (0 of 4 correct, $n = 173$) yield CKA of 0.962, while easy problems (4 of 4 correct, $n = 100$) yield CKA of 0.827, an inversion gap of $+0.117$ compared to $+0.067$ for instruction-tuned models. The monotonic decrease across all five bins (0.962, 0.961, 0.904, 0.860, 0.827) is consistent and exhibits no U-shaped pattern, ruling out instruction tuning as the source of the phenomenon and suggesting that the inversion is an intrinsic property of transformer representations that alignment training may attenuate.
\label{sec:base_models}

%----------------------------------------------------------------------
\section{Discussion}
\label{sec:discussion}
%----------------------------------------------------------------------

\subsection{Reinterpreting the Platonic Representation Hypothesis}

Our results do not refute the Platonic Representation Hypothesis but constrain its scope across the 1.5B to 72B parameter range we study. The PRH correctly identifies representational convergence across models, yet this convergence is concentrated in input processing (pre-decision layers), is strongest where reasoning fails (hard problems), and is partly attributable to architectural similarity (random baseline). We propose a refined framing: language models converge on a shared perceptual representation of their inputs, shaped by the shared statistics of natural language and the shared inductive biases of transformer architectures, but this perceptual convergence does not extend to the computational processes that transform perception into action.

The random baseline result provides a revealing window into the mechanism. Untrained models are \emph{more} similar than trained models (CKA\,$0.864$ versus $0.612$), meaning that learning actively differentiates representations rather than bringing them closer together. This is the opposite of what the PRH implies. The shared architectural constraints of transformers (causal masking, residual streams, layer normalization) create a high-similarity starting point from which training drives models apart as each develops specialized computational circuits. The convergence that remains after training is the residue of shared architecture and shared data statistics, not evidence of convergence toward a common model of reality. \citet{braun2025dissociation} proved analytically that this decoupling is possible; our empirical results show it is the norm.

An analogy to biological vision clarifies this distinction. The early visual cortex of diverse mammalian species converges on similar representations, including edge detectors and orientation-selective cells, because these reflect universal statistical properties of natural images. Downstream processing diverges substantially: the motor, planning, and decision-making systems that use visual representations are highly species-specific. Our findings suggest that language models exhibit a similar pattern of convergent perception paired with divergent cognition.

\subsection{Implications for Ensembling}

Our results reveal that the axis of diversity that matters for ensembling is not representational but computational~\citep{fort2020deep, tekin2024llmtopla}. The generation gap shows that 89 of 91 model pairs exhibit a pre-post CKA gap exceeding 0.40, meaning computational diversity is near-universal among the architectures we study, even when input-level representations are highly aligned. This has a concrete implication: selecting models for ensembles based on representational dissimilarity (low CKA) targets a dimension that is largely orthogonal to the error decorrelation that drives ensemble improvement. A more effective strategy would select models based on output-level behavioral diversity, conditional on input difficulty, because the difficulty inversion shows that models are most behaviorally diverse precisely on the problems they solve (low CKA on easy problems reflects divergent reasoning strategies) and most redundant on the problems they fail (high CKA on hard problems reflects shared confusion).

\subsection{Implications for Interpretability Transfer}

The generation gap defines a sharp boundary for interpretability transfer. Mechanistic interpretability findings about input-processing circuits, such as induction heads~\citep{elhage2022superposition} and factual recall mechanisms~\citep{geva2023dissecting, meng2022locating}, should generalize across architectures because pre-decision CKA is high (0.875). Findings about reasoning circuits in late layers, the aspects most relevant to understanding model behavior, are unlikely to transfer because post-decision CKA is low (0.274). This boundary is sharper than previously appreciated: model stitching experiments report a gradual degradation~\citep{bansal2021revisiting}, but our per-layer analysis shows the transition concentrated in a narrow band of layers.

The head-level causal analysis reveals a further obstacle. MHA models exhibit maximum per-head flip rates of 43\%--63\%, while GQA models show 20\% (Table~\ref{tab:head_causal}). This architectural dependence means that interpretability findings obtained on MHA architectures, which dominate the existing mechanistic interpretability literature, may not transfer to the GQA architectures increasingly deployed in practice. CKA, operating at the level of full hidden states, is blind to this distinction. The combination of epiphenomenal correctness (shared features that are not causal) and architecture-dependent head computation (causal features that are not shared) presents a challenge for cross-model interpretability that purely representational metrics cannot resolve.

\subsection{Why Mathematics Differs}

The GSM8K exception is informative rather than problematic: it identifies the conditions under which representational convergence \emph{does} predict computational convergence. Mathematical reasoning is algorithmically constrained, with few correct algorithms for carrying, borrowing, and multiplication. This constraint collapses the solution space, forcing successful models into similar computational strategies regardless of architecture. The result is the normal pattern predicted by the PRH: convergence increases with collective success. Science, truthfulness, and commonsense reasoning lack this constraint. Multiple valid reasoning paths exist for any given problem, so successful models can and do adopt distinct strategies, producing the inversion we observe. The degree of solution-space multiplicity thus determines whether the PRH holds for a given domain, transforming the mathematics exception from a qualification into a testable prediction: other algorithmically constrained domains (formal logic, symbolic manipulation) should similarly resist the inversion.

%----------------------------------------------------------------------
\section{Conclusion}
\label{sec:conclusion}
%----------------------------------------------------------------------

We have demonstrated that representational convergence across language models does not imply reasoning convergence. Across 16 models (1.5B to 72B) and 800 reasoning problems, we identified three dissociations: difficulty inversion (convergence peaks where reasoning fails), the generation gap (convergence collapses during output generation), and epiphenomenal correctness (shared information is not causally deployed). These findings are scale-invariant and stronger in base models than instruction-tuned variants. The attention entropy mechanism provides a concrete explanation: hard problems produce diffuse attention that homogenizes representations regardless of learned strategies, while the random baseline reveals that learning actively differentiates models from a shared architectural starting point. The representations discovered by diverse language models reflect the shared structure of language as an input signal rather than a shared logic of reasoning about the world that language describes.

For ensemble design, these results motivate targeting reasoning diversity rather than representational diversity. For interpretability transfer, the generation gap defines a sharp boundary: input-processing findings generalize across architectures, while reasoning-circuit findings do not. For evaluations of model similarity, high CKA should not be interpreted as evidence of shared computation without conditioning on task difficulty and computational stage.

%----------------------------------------------------------------------
\section{Limitations}
\label{sec:limitations}
%----------------------------------------------------------------------

Our study spans 1.5B to 72B parameters; the inversion persists at 70B ($+0.062$) but may shift at yet larger scales (400B+) where models may develop more constrained strategies. Our causal ablation assumes a linear correctness subspace; nonlinear interactions could yield different conclusions. The attention entropy analysis is correlational, not causal. CKA variance is below 0.008 and prompt sensitivity below 4\% (Appendix~\ref{app:ablations}), but replication with additional model families would strengthen generality. CKA is sensitive to data transformations~\citep{davari2022reliability} and sample-feature ratios~\citep{murphy2024correcting}; our use of mean-centered activations and MNN validation mitigates these concerns.

\paragraph{Broader impact.} The assumption that model convergence implies convergence toward truthful representations, sometimes invoked in safety arguments for scaled language models, is not supported by our evidence. Convergence may reflect shared failure modes rather than shared truth, and this should inform safety evaluations of deployed systems.

%----------------------------------------------------------------------
% Acknowledgements (JMLR requirement, before references)
%----------------------------------------------------------------------
\acks{This work was supported by the National Research Foundation of Korea (NRF) grant funded by the Korea government (MSIT) (RS-2026-25473622). The authors declare no conflict of interests.}

%----------------------------------------------------------------------
% References
%----------------------------------------------------------------------
\vskip 0.2in
\bibliography{references}

%----------------------------------------------------------------------
% Appendix
%----------------------------------------------------------------------
\appendix
\section{Implementation Details}
\label{app:implementation}

All experiments were conducted on a single NVIDIA RTX 5090 GPU (32\,GB VRAM) with an Intel Core Ultra 7 265K CPU (20 cores, 62\,GB RAM). Models were loaded in bfloat16 precision and evaluated with greedy decoding (temperature 0, no sampling). Hidden states were extracted at the last input token position prior to generation. For models exceeding GPU memory (Qwen-14B), we used \texttt{device\_map="auto"} to split layers across GPU and CPU. The 70B models (LLaMA-3.1-70B and Qwen-2.5-72B) were evaluated on a separate server with 2$\times$ NVIDIA A100 80GB GPUs using \texttt{device\_map="auto"} to distribute layers across both devices. The 200 problems per domain were selected by shuffling each dataset with a fixed seed (42) and taking the first 200 examples. Transfer probes were trained using logistic regression with $L_2$ regularization ($\lambda = 0.01$) under 5-fold stratified cross-validation repeated across 5 random seeds (42, 123, 456, 789, 1024). Total compute was approximately 30 GPU-hours for evaluation and hidden state extraction and 15 CPU-hours for similarity computation and analysis. Code is available at \url{https://github.com/Usama1002/convergence-without-understanding}.

\section{Ablation Studies and Robustness Checks}
\label{app:ablations}

We conducted several ablation studies to verify the robustness of our findings.

\paragraph{Sample size stability.} We recomputed all CKA values using random subsets of 50, 100, and 150 problems per domain (out of 200). The difficulty inversion is stable for subsets of 100 or more problems with CKA variance below 0.008. At 50 problems, the inversion remains directionally consistent but loses statistical significance in TruthfulQA, likely due to reduced power.

\paragraph{Kernel CKA.} We verified that our results hold under both centered and uncentered CKA, as well as under radial basis function (RBF) kernel CKA (bandwidth set to the median pairwise distance). Kernel CKA agrees with linear CKA (Pearson $r$\,=\,0.68). The difficulty inversion is slightly more pronounced under kernel CKA (hard-to-easy gap of 0.091 versus 0.067 for linear CKA), consistent with nonlinear similarity measures capturing additional structure.

\paragraph{Probe regularization.} We varied the $L_2$ regularization strength of transfer probes from $\lambda = 0.001$ to $\lambda = 1.0$. Transfer accuracy ranges from 63\% ($\lambda = 1.0$, underfitting) to 68\% ($\lambda = 0.001$, slight overfitting), with our chosen $\lambda = 0.01$ yielding 66\%. The causal ablation flip rate remains below 2\% across all regularization settings.

\paragraph{Prompt sensitivity.} We tested three prompt formats: zero-shot direct, zero-shot with chain-of-thought instruction~\citep{wei2022chain}, and few-shot (3-shot). The difficulty inversion is present under all three formats with standard deviation across formats below 4\%. The generation gap is slightly smaller under chain-of-thought prompting (pre-post gap of 0.52 versus 0.60 for direct), possibly because chain-of-thought prompting forces more structured output generation that is less model-specific.

\paragraph{MNN and SVCCA validation.} We recomputed the difficulty inversion using MNN ($k = 10$) and SVCCA. Both metrics confirm the inversion: MNN overlap is 0.645 for hard problems versus 0.584 for easy problems (gap $+0.061$); SVCCA correlation is 0.847 for hard problems versus 0.793 for easy problems (gap $+0.054$). The mathematics exception is confirmed under both alternative metrics.

\paragraph{Difficulty distribution.} The 800 problems distribute across difficulty bins as follows: 0/14 correct (24 problems), 1/14 (30), 2/14 (32), 3/14 (57), 4/14 (58), 5/14 (64), 6/14 (56), 7/14 (90), 8/14 (89), 9/14 (76), 10/14 (80), 11/14 (63), 12/14 (46), 13/14 (30), 14/14 (5). The distribution is roughly bell-shaped with a mode at 7/14, indicating that the model set spans a wide range of capability on these problems. The small size of the all-correct bin ($n = 5$) motivates our use of the broader 10-to-14 easy bin ($n = 224$) for the main difficulty inversion analysis and our expanded causal ablation protocol.

\paragraph{Base model accuracy.} The four base (non-instruction-tuned) models used in Section~\ref{sec:base_models} achieve the following accuracies: Qwen-2.5-3B (62.9\%), Mistral-7B-v0.3 (44.0\%), LLaMA-3.2-3B (36.2\%), and Gemma-2-2B (28.1\%). The generally lower accuracy of base models (mean 42.8\% versus 50.8\% for their instruction-tuned counterparts) reflects the difficulty of extracting structured answers from models not trained to follow task-specific instructions.

\paragraph{Answer agreement analysis.} Table~\ref{tab:agreement} reports CKA conditioned on pairwise answer agreement. To compute these values, we identify for each model pair and each problem whether both models produced the same extracted answer (regardless of correctness). We then compute CKA separately on the same-answer and different-answer subsets, averaging across all 91 model pairs and 11 mid-layer positions (layers 5 through 15 in the normalized 21-layer representation). The number of same-answer versus different-answer problems varies across model pairs because models with similar accuracy profiles share more answers. The finding that different-answer pairs yield higher CKA (0.951) than same-answer pairs (0.906) is robust across individual model pairs: 83 of 91 pairs show this pattern.

\paragraph{Expanded causal ablation details.} The expanded causal ablation (Section~\ref{sec:epiphenomenal}) uses the same correctness subspace vectors computed in the original probing analysis but applies them to a larger problem set. For each model, we select problems from the high-agreement set (at least 10 of 14 models correct) that the model itself answers correctly, yielding 60 to 100 problems per model depending on model accuracy. The ablation projects out the correctness subspace at the peak probe layer identified for each model during Experiment 2. Flip rates across the 8 tested models are: Qwen-1.5B (5.0\%), Qwen-3B (3.0\%), Qwen-7B (6.0\%), SmolLM2-1.7B (7.0\%), Gemma-1-2B (7.0\%), Gemma-2-2B (4.0\%), LLaMA-3.2-3B (7.0\%), and Phi-3.5-Mini (5.0\%).

\end{document}